\documentclass[10pt,twocolumn,letterpaper]{article}

\usepackage{cvpr}
\usepackage{times}
\usepackage{epsfig}
\usepackage{graphicx}
\usepackage{amsmath}
\usepackage{amssymb}

\usepackage[breaklinks=true,bookmarks=false]{hyperref}

\cvprfinalcopy 

\def\cvprPaperID{****} 

\ifcvprfinal\fi
\begin{document}

\title{Interpretable 3D Human Action Analysis with Temporal Convolutional Networks }

\author{Tae Soo Kim\\
Johns Hopkins University\\
Baltimore, MD\\
{\tt\small tkim60@jhu.edu}
\and
Austin Reiter\\
Johns Hopkins University\\
Baltimore, MD\\
{\tt\small areiter@cs.jhu.edu}
}

\maketitle

\begin{abstract}
   The discriminative power of modern deep learning models for 3D human action recognition is growing ever so potent. In conjunction with the recent resurgence of 3D human action representation with 3D skeletons, the quality and the pace of recent progress have been significant. However, the inner workings of state-of-the-art learning based methods in 3D human action recognition still remain mostly black-box. In this work, we propose to use a new class of models known as Temporal Convolutional Neural Networks (TCN) for 3D human action recognition. Compared to popular LSTM-based Recurrent Neural Network models, given interpretable input such as 3D skeletons, TCN provides us a way to explicitly learn readily interpretable spatio-temporal representations for 3D human action recognition. We provide our strategy in re-designing the TCN with interpretability in mind and how such characteristics of the model is leveraged to construct a powerful 3D activity recognition method. Through this work, we wish to take a step towards a spatio-temporal model that is easier to understand, explain and interpret. The resulting model, Res-TCN, achieves state-of-the-art results on the largest 3D human action recognition dataset, NTU-RGBD.
\end{abstract}

\section{Introduction}

Human activity analysis is a crucial yet challenging research area of computer vision. Applications of human activity recognition ranges from video surveillance, human-computer interaction, robotics and skill evaluation \cite{survey2014,survey2011}. At the core of successful systems for human activity recognition lies an effective representation that can model both the spatial and temporal dynamics of human motion. 

Traditionally, the community has focused on activity recognition in the domain of RGB videos \cite{survey2008,survey2016}. For a RGB video, complex human motion in 3D euclidean space is projected on to a series of 2D images and in the process, loss of valuable 3D spatio-temporal information is inevitable. In recent years, we have witnessed a drastic improvement of cost-effective depth sensors in the form of Microsoft Kinect \cite{kinect_review2013}. Naturally, computer vision methods leveraging on the 3D structure provided by such 3D sensors, namely RGB+D methods, have been an active area of research \cite{skeleton_review_2017}. Applied to human activity recognition, 3D information of how a human body articulates comes in the form of time series sequence of 3D skeletons. Such representations describe human motion as a collection of trajectories in 3D euclidean space of key human joints. Even without the context information and visual cues, early work \cite{Johansson1973} in biological perception and more recent methods \cite{skeleton_review_2017,kinect_review2013,Song2017} provide strong evidence that encoding humans as a 3D skeleton yields both a discriminative and a robust representation for activity analysis. Given the recent progress of powerful human pose estimators from RGB or RGBD data \cite{pose_estimation_2016}, human activity recognition model that builds on top of 3D skeletons is a promising direction.

Despite this significant progress, the inner workings of such complex temporal models still remain mostly black-boxes. Without the capability to interpret learning based models, we inevitably lack the power to fully support a model's decision regardless of its correctness \cite{mythos,trust2016}. Such short-comings may hinder practical deployment of even the strongest models. The ability to understand and explain precisely how a model came to a wrong prediction is a fundamental first step towards improving the potential of our current methods.

In this light, we propose Temporal Convolutional Neural Networks (TCN) \cite{TCN} applied to 3D Human Action Recognition. Through the lens of TCN, we wish to uncover what exactly learning-based temporal models leverage on especially when trained on interpretable data such as a sequence of 3D skeletons. We re-design the original TCN by factoring out the deeper layers into additive residual terms which yields both interpretable hidden representations and model parameters. Using the resulting architecture, Res-TCN, we validate our approach on currently the largest 3D human activity recognition dataset, NTU-RGBD and obtain state-of-the-art results.

\section{Related Work}
In this section, we first provide a literature review on recent developments in learning based 3D human action recognition models. We focus our narrative on models that employ LSTM-based Recurrent Neural Networks. We also extend our review to works focusing on model interpret-ability and  visualization of deep learning models.

\subsection{3D Human Activity Recognition}
Traditional recurrent neural network models suffer from vanishing/exploding gradient problem during optimization and are difficult to train correctly \cite{difficulty,bilstm}. By formulation, LSTM neurons begin to address such optimization problems and are capable of modeling long-term dependencies \cite{lstm,lstm2,bilstm}. Given the temporal recurrent nature of human action analysis, most leading methods in 3D human action recognition adopt LSTM-based RNNs. 

Hierarchical recurrent neural network of \cite{du2015} combines the features of different body parts hierarchically. At the initial layer, each sub-network extracts features over a single joint and these representations are fused hierarchically in the deeper layers. A final prediction is made when all joint information is combined. In part-aware LSTM model introduced in \cite{nturgdb2016}, individual body joints are grouped together in five groups based on their spatial context. The memory units of the LSTM are learned independently per group and the information from different parts is aggregated to produce a final prediction. The work of \cite{zhu2016} leverages on similar intuition that co-occurrence of joints is a strong discriminative feature for human action recognition. A group sparsity constraint on the connection matrix pushes the network to learn the mappings between co-occurring joints and the human activity. Deep spatio-temporal LSTM with Trust Gates is introduced in \cite{Liu2016} to learn features both in the temporal and spatial domains. Similarly, the authors of \cite{Song2017} propose a spatio-temporal attention model for LSTM-based RNNs. The method comprises of three LSTM networks: a spatial attention sub-LSTM, a temporal attention sub-LSTM and a main LSTM. Both temporal and spatial attention modules are pre-trained separately initially and the entire network is trained end-to-end. 

In the above methods, the key intuition is that a certain subset of joints are more important for recognizing human activities. However, it is difficult to interpret what the model parameters of each LSTM layer represent. In our proposed version of  TCN, we show that our model also learns both spatial and temporal attention without the need for initial pre-training stage as in \cite{Song2017}. Moreover, by model design based on temporal convolutions \cite{TCN} and residual connections \cite{resnet}, we can begin to directly interpret what our model parameters and features represent. 

\subsection{Model Interpretability and Visualization}
Here, we focus our discussion on interpretability of supervised machine learning models. Post-hoc approaches are often considered to provide interpretation of models. This means that once a model is learned, post-operative experiments are conducted to gain insight into what the model has learned. \cite{high-layer} proposes a method to find the optimal stimulus for each unit in a deep neural network by performing back propagation with respect to image space to maximally stimulate a neuron. Obtained images give us an insight into the appearance of the input that a neuron is most likely to activate. The work of \cite{Simonyan2013} sheds light on what spatial context of the image the convolutional neural network (CNN) is leveraging on for image classification through saliency maps. Similarly, the authors of \cite{Zeiler2014} use a deconvnet \cite{deconv} to map the activities of intermediate layers back to the input pixel space so that inputs that maximally activates an intermediate layer can be directly visualized as an image. The methods mentioned above uncover that CNNs learn to decompose the image space into hierarchical modular patterns. However, not all visualized patterns are necessarily interpretable or understandable. In such post-hoc approaches, there is no control over how the model is optimized in the first place. Though it is valuable and interesting to expose what the model has learned after-the-fact, we wish to take a more active approach to the problem. We focus our investigation on how to improve model interpretability by design. 

Another popular direction in post-hoc approaches is the use of examples and prototypes. Example-based explanations and classifiers have shown to offer a condensed view of a dataset, potentially offering a reason why a classifier came to a certain conclusion through other data points in the dataset \cite{prototype}. The work of \cite{critic} pushes this idea further by forcing a model to produce both exemplars and criticisms. Even with such examples, the causality between model parameters and the final prediction of the model is still unclear. In our work, we strive to take a more direct approach on model interpretability. We focus on two key questions: 1. How do we interpret the representations learned using TCN and 2. How can we design a deep learning architecture that provides readily interpretable hidden representations and model parameters in the context of 3D human action analysis?

\begin{figure*}
\begin{center}
\includegraphics[width=0.90\textwidth,height=7cm]{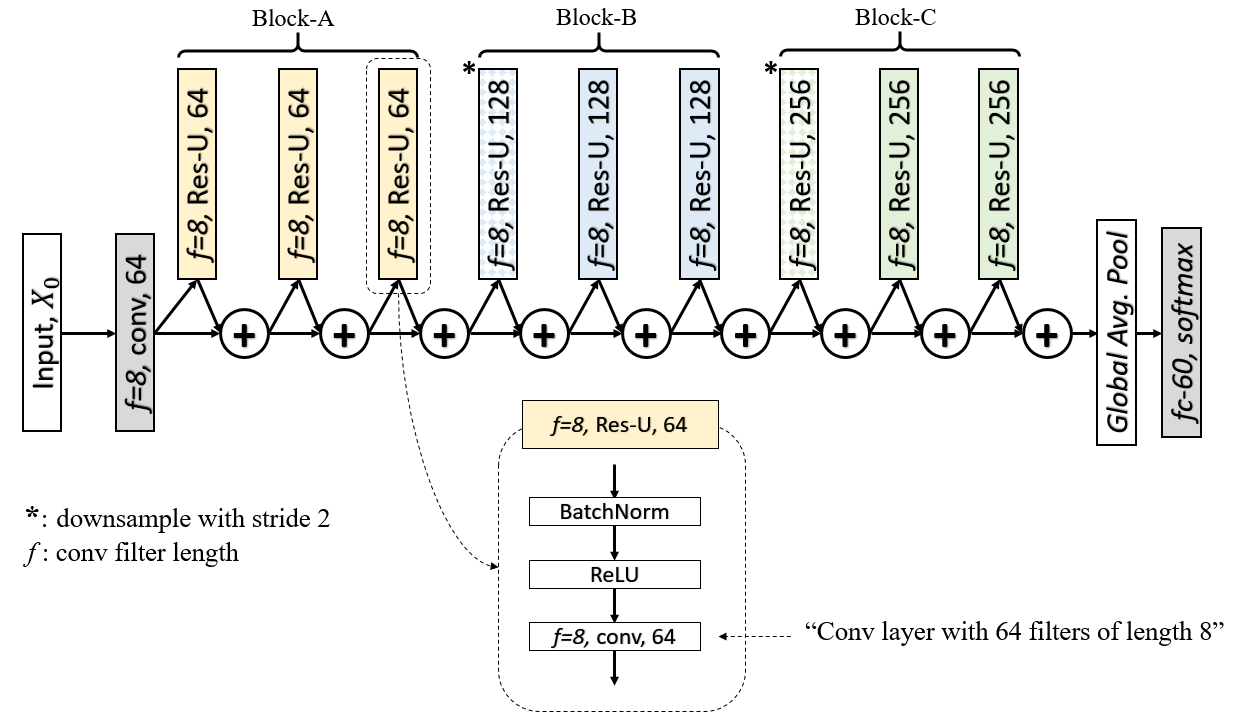}
\end{center}
   \caption{Res-TCN model architecture. Except the first convolution layer (in gray), the model consists of stacked residual units.}
\label{fig:model}
\end{figure*}

\section{Overview of Temporal Convolutional Neural Networks}
In this section, we provide a brief overview of the structure of a TCN as provided in the original paper \cite{TCN}. Note that the original TCN is designed for temporal action segmentation in video and it follows a convolutional encoder-decoder design. We adapt the encoder portion of the net for action recognition. The properties of a TCN follow those of a modern spatial Convolutional Neural Network (CNN) for recognition \cite{lecun, alexnet, vgg} and segmentation tasks  \cite{segnet}. The network is built from stacked units of 1-dimensional convolution followed by a non-linear activation function. The 1-dimensional convolution is across the temporal domain.

The input to an original TCN is a sequence of video features. A $D$-dimensional feature vector, whether it is a deep feature from a spatial CNN such as $fc7$ activation of AlexNet \cite{alexnet} or a set of kinematic features \cite{TCN}, is extracted per each video frame. For a video of $T$ frames, the input $X$ is simply a concatenation of all frame-wise features such that $X \in \mathbb{R}^{T \times D}$. 

As with well-known CNN models, repeated blocks of convolutions followed by non-linear activations extract features from the input. More precisely, in a TCN, the $l$-th convolution layer with a temporal window of $d_l$ consists of $N_l$ filters $\{W^{(i)}\}_{i=1}^{N_l}$ where each filter is $W^{(i)} \in \mathbb{R}^{f_l \times N_{l-1}}$. Given an output $X_{l-1}$ of the previous layer, the $l$-th layer output, $X_l$ is simply
\begin{equation} \label{eq:1}
 X_l = f(W*X_{l-1})
\end{equation}
where $f$ is a non-linear activation function such as ReLU. The whole network is trained with back-propagation. In an attempt to further improve the interpretability of a TCN, we adopt residual connections of \cite{resnet}. In the following sections, we discuss how such skip connections and the resulting TCN architecture, namely Res-TCN, leads to improved interpretability of 3D human action recognition models. 

\section{Interpretability of TCNs with Residual Connections}

In our work, for the purpose of 3D human activity recognition, the input to the model $X_0$ is a frame-wise skeleton features concatenated temporally across the entire video sequence. An important pre-requisite of the interpretability of TCNs is that each dimension of the frame-wise feature must be interpretable as well. Let $x_t$ be a skeleton feature extracted from a video frame $t$ where $x_t \in \mathbb{R}^{D}$. By construction of the skeleton feature, the $d$-th dimension of the feature, $x_t(d)$, has an interpretable meaning associated with it (for example, Z position of the right elbow joint). The details of feature construction will be covered in the experiments section.

\begin{figure*}
\begin{center}
 \includegraphics[height=7.5cm,keepaspectratio]{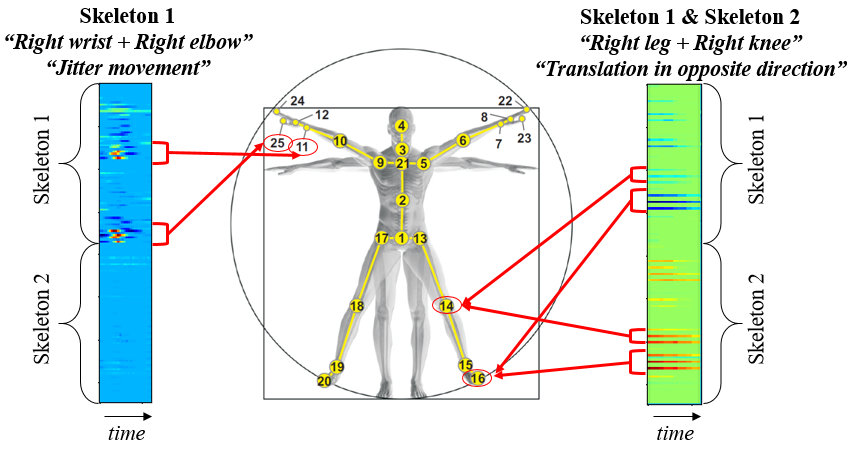}
\end{center}
 \caption{Examples of direct mapping from layer 1 filter parameters to skeleton joints.}
\label{fig:filter}
\end{figure*}

\subsection{TCN with Residual Units and Identity Mappings}
The biggest road-block in interpreting current spatio-temporal models such as LSTM-based RNNs for 3D human action analysis stems from the lack of clear connection between the learned model parameters and their hidden representations. However, for TCNs, the formulation of hidden representations from its model parameters is straightforward to comprehend: activation maps are computed by convolving a learnable temporal filter across time and passing the output through a ReLU unit. In a ReLU network, after an iteration of a forward-backward pass, the network parameters are optimized such that convolution of a filter across the characteristic regions of the input more likely produces a \textit{positive} value in the next iteration. We can exploit such behavior of the model to improve the model interpretability by re-formulating the TCN with residual connections \cite{resnet}.

As introduced in \cite{resnet}, skip connection with identity mapping introduces beneficial properties for network convergence even for very deep networks. We observe that such design for CNNs improves model interpretability as well given input with semantic meaning. Our Res-TCN model architecture is shown in Figure \ref{fig:model}. 

Res-TCN stacks building blocks called \textit{Residual Units} as introduced in \cite{resnet1} and adapts the pre-activation scheme of \cite{resnet}. Each unit in layer $l$ performs the following computation:

\begin{equation} \label{eq:2}
 X_{l} = X_{l-1} + F(W_l,X_{l-1})
\end{equation}

\begin{equation} \label{eq:3}
F(W_l,X_{l-1}) = W_l*\sigma(X_{l-1})
\end{equation}

$F$ denotes the residual unit. For the $l$-th layer, $X_{l-1}$ denotes the input, $W_l$ is the set of learnable parameters and $\sigma$ is a ReLU activation function. We can re-write the expression $W_l*\sigma(X_{l-1})$ as $W_l*max(0,X_{l-1})$ when $\sigma$ is ReLU. The only exception in our architecture is the very first convolution layer. The first convolution layer in Res-TCN operates on raw skeleton input and the resulting activation map, $X_1$, is passed on the subsequent layers. Given a Res-TCN with $N$ residual units, the hidden representation after $N$ residual units is:

\begin{equation} \label{eq:4}
X_N = X_1 + \sum_{i=2}^{N} W_i*max(0,X_{i-1})
\end{equation}

\begin{equation} \label{eq:5}
X_1 = W_1X_0
\end{equation}

Note that $X_1$ is a result of convolving a set of filters in layer $l=1$ without undergoing any non-linear activation. The set of filters in $W_1$ and the resulting activation map, $X_1$, are directly interpretable given that each dimension of $X_0$ is directly interpretable as well, such is the case when $X_0$ is a set of skeleton features. An important observation in our design is that in the $l$-th residual unit where $l \geq 2$, ReLU is performed on $X_{l-1}$ prior to applying convolution with filters in $W_l$. In other words, the gradient only flows through the positive regions of $X_{l-1}$ and $W_l$ learns to pick up discriminative patterns where $X_{l-1} > 0$. The computation $W_l*max(0,X_{l-1})$ is then added to the input, $X_{l-1}$, and passed on to the next layer. The input to the first residual unit is $X_1$ and all subsequent residual units in a Res-TCN either adds to or subtracts from $X_1$ as shown in Equation \ref{eq:4}. In this formulation, we are forcing the network to learn discriminative spatio-temporal features in the \textit{common language} of $X_1$. In the experiments section, we visualize hidden representation of an activity from one of the deeper layers and show its connection to $X_1$ to validate our analysis.

For prediction, we apply global average pooling after the last merge layer across the entire temporal sequence and attach a softmax layer with number of neurons equal to number of classes.

\subsection{A Closer Look at Model Parameters}
\label{section:closer}
In a Res-TCN architecture, Equation \ref{eq:4} suggests that the representational power of the entire model depends heavily on producing discriminative $X_1$ through filters in $W_1$. In this section, we analyze what each filter in $W_1$ represents. 

Consider a single 1D convolution filter $W^{(k)}_1$ from $\{W^{(i)}_1\}_{i=1}^{N_1}$. $W^{(k)}_1$ computes 1D convolution over $X_0 \in \mathbb{R}^{T \times D}$ with a defined stride of $s$ and a filter length of $f_1$. Examples of converged filters are shown in Figure \ref{fig:filter}.

Each filter looks at $f_1$ time steps concurrently across all feature dimensions such that $W^{(k)}_1 \in \mathbb{R}^{f_1 \times D}$. An important property of $W_1$ filters is that each dimension $d \in {1,\ ...\ ,D}$ has an explainable meaning associated with it. For example, the $d$-th dimension of a skeleton feature $x_t(d)$ depicts a spatial configuration (euclidean X, Y or Z coordinate wrt. the depth sensor) of a particular joint at time $t$. For instance, the filter depicted on the left in Figure \ref{fig:filter} has parameters close to zero for all joint location except at indices corresponding to joint numbers 11 and 25 as defined in \cite{nturgdb2016}. We can take a step further in interpreting this filter: the filter directly encodes how the joints move through time. In a time window defined by $f_l$, for filter weights associated with joint indices 11 and 25, the weights increase sharply towards a peak and then returns back to their starting magnitudes. We know that in order for this particular filter to produce a high positive convolution score, the input $X_0$ at corresponding dimensions must exhibit a highly correlated sequence structure to that of the filter. We can then provide a clear explanation of what this filter is looking for: "A quick jitter movement of the joints in the right hand". Similarly, consider the filter on the right in Figure \ref{fig:filter}. By design, the bottom half of the parameters correspond to the second actor. Following the same logic described above, we can clearly understand that this filter produces a high positive convolution score with an input where two actors are translating apart from each other. In the following section, we extend our discussion to parameter interpretability in the deeper layers.

\begin{figure}[t]
\begin{center}
 \includegraphics[width=8.75cm]{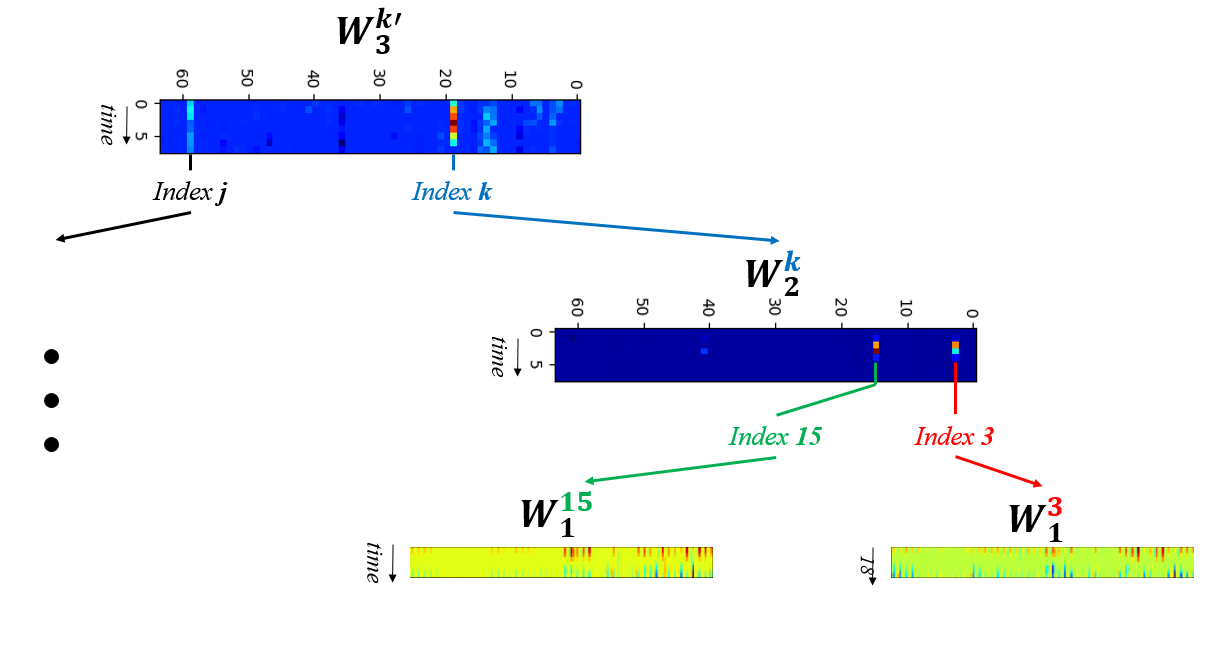}
\end{center}
 \caption{An example how we can map convolution filters in deeper layers, such as $W_3^{k'}$, all the way back to filters in the first layer $W_1$.}
\label{fig:deep_filter}
\end{figure}

\subsection{A Deeper Look at Model Parameters}
\label{section:deeper}
Let us now extend our analysis to deeper layers in the model. In a Res-TCN formulation, deeper layers are factored out into residual units and an output from a residual unit is simply merged by adding to the input of the residual unit. For example, consider the hidden representation after two convolution layers:

\begin{equation} \label{eq:6}
X_2 = X_1 + W_2*max(0,X_1)
\end{equation}
where $X_1 = W_1*X_0$. Filters in $W_2$ convolve over the positive regions of the output produced by $W_1*X_0$  such that $W^{(k)}_2 \in \mathbb{R}^{f_2 \times N_1}$ for some $k \in [0,N_2)$ where $f_2$ defines the filter length in layer $l=2$, $N_1$ is the number of filters in layer $l=1$ and likewise for $N_2$. Following the formulation of Equation \ref{eq:6}, we observe that $W_2$ acts as a gate that modulates how much information will be transformed and added on to $X_1$. Given a filter $W^{(k)}_2$, a large weight value in the $d$-th dimension of $W^{(k)}_2$ indicates that this particular filter adds to or subtracts from $X_1$ a weighted version of the incoming signal at the same dimension $d$ where $d \in [0,N_1]$. Dimensions with low weight magnitudes contributes less to the final output of the residual unit. Consider example filters from deeper convolution layers shown in Figure \ref{fig:deep_filter}. Most parameters are close to zero except in certain dimensions. Given the additive nature of residual units, we can directly map such dimensions to filters in the lower layer. If dimension $k$ of $W_3^{(k')}$ has high weight magnitudes, then $W_3^{(k')}$ allows more information computed from $W_2^{(k)}$ to be added to the output. We can recursively trace down such \textit{influential} filters all the way down to the very first convolution layer where we can directly map filter parameters to interpretable skeleton motion as shown in the previous section. 

\begin{figure*}
\begin{center}
 \includegraphics[height=7.5cm,keepaspectratio]{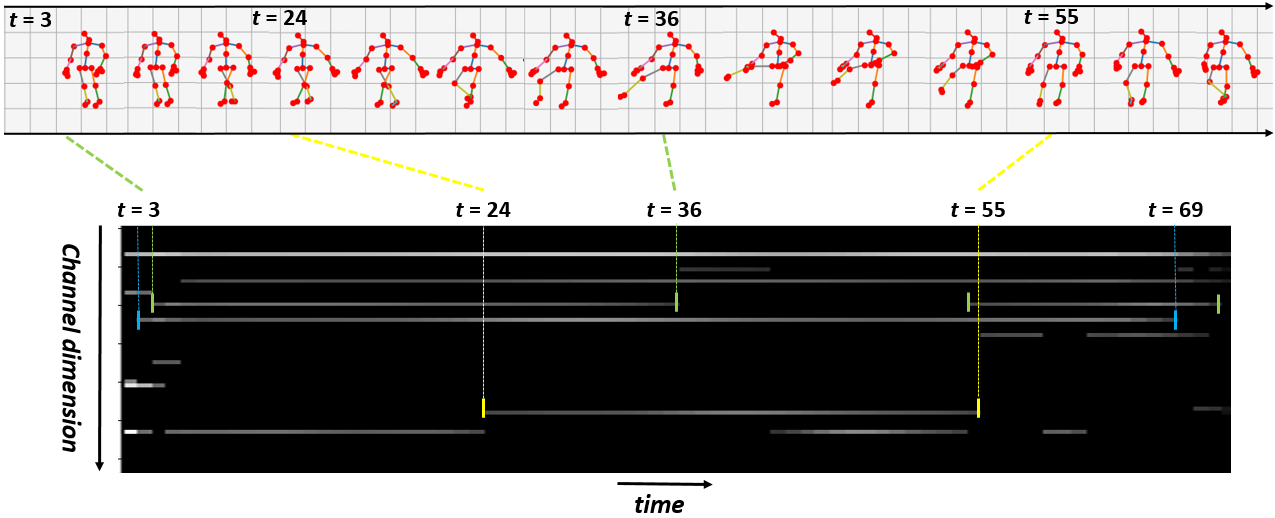}
\end{center}
 \caption{An example skeleton sequence, its hidden representation ($X_4$) in Res-TCN and associated filters from $W_1$.}
\label{fig:exp_vis}
\end{figure*}

\section{Experiments}
We validate that our approach not only leads to an interpretable representation but also to a discriminative one. We evaluate Res-TCN on 3D skeleton based human activity recognition dataset of NTU \cite{nturgdb2016}. We also provide interpretations on our model predictions based on the concepts discussed in the previous sections.

\subsection{Dataset and Settings}
NTU RGB+D dataset \cite{nturgdb2016} is currently the largest human activity recognition dataset with full 3D skeleton annotations. It contains 56880 training videos ranging over 60 action classes. The dataset provides two train/test split paradigms: Cross-Subject (CS) and Cross-View (CV) settings. The dataset covers 40 distinct subjects with varying physical traits. In terms of camera viewpoints, three cameras are placed in three different angles:$-45^{\circ}$, $0^{\circ}$ and $+45^{\circ}$.

\textbf{Implementation Details}: We follow the skeleton feature construction procedure as adapted in \cite{nturgdb2016,Song2017}. However, in contrast to their feature extraction stage, we do not perform view normalization prior to feeding the features into our Res-TCN. We take the raw (X,Y,Z) values of each skeleton joint and concatenate all values to form a skeleton feature per frame. Given that there are at most two actors in the scene and there are 25 joints per skeleton, a skeleton feature per frame is a $2*25*3=150$ dimensional vector. We use the Keras deep learning framework \cite{keras} with a TensorFlow backend \cite{tensorflow}. We use an initial learning rate of 0.01 and decrease the learning rate by a factor of 10 when the testing loss plateaus for more than 10 epochs. We use stochastic gradient descent with nesterov acceleration with a momentum of 0.9. L-1 regularizer with a weight of $1e^{-4}$ is applied to all convolution layers. We use a batch size of 128. Dropout \cite{dropout} with rate 0.5 is applied after all activation layers to prevent overfitting. We perform all our experiments on a Nvidia K80 GPU. The implementation and converged model weights will be made publicly available \footnote{https://github.com/TaeSoo-Kim/TCNActionRecognition/}.

\subsection{Why Did My Model Predict This?}
Leveraging on the explainable structure of Res-TCN, we wish to provide an answer to the question: "How/why did my model come to this conclusion?", using only the model parameters and hidden representations as the basis for providing such an explanation. 

Let us choose an arbitrary video clip from NTURGB+D. The particular sequence of skeletons that we visualize in Figure \ref{fig:exp_vis} is of class \textit{kicking something} and is approximately 70 frames long. The output of the first block (Block-A in Figure \ref{fig:model}) is displayed above. As discussed in section \ref{section:deeper}, we can trace which of the $W_1$ filters had the largest influence on any given deeper hidden representation $X_n$ where $n > 1$. For clarity in visualization, for each time step, we only plot the activation values that are within the top 20 percentile. Each column denotes the activation values from all filters in $W_4$ and each row denotes the corresponding filter's response over time. Consider the dimension in the activation map that is color coded with green in Figure \ref{fig:exp_vis}. By following the logic described in section \ref{section:closer},  we found that this particular filter produces a high positive response for translational movement of the left ankle and left hip. The yellow filter has high magnitude parameters associated with the right knee joint. And finally, the blue filter picks up signals from the right ankle and the left wrist joint. The activation map of $X_4$ and the corresponding $W_1$ filters tell a rather detailed and precisely timed story about the input skeleton sequence: \textit{the left ankle and hip joints first translate followed by a sudden movement of the right knee, all the while the left wrist and the right ankle undergo a swinging motion}. 

\begin{figure}[t]
\begin{center}
 \includegraphics[width=8.75cm]{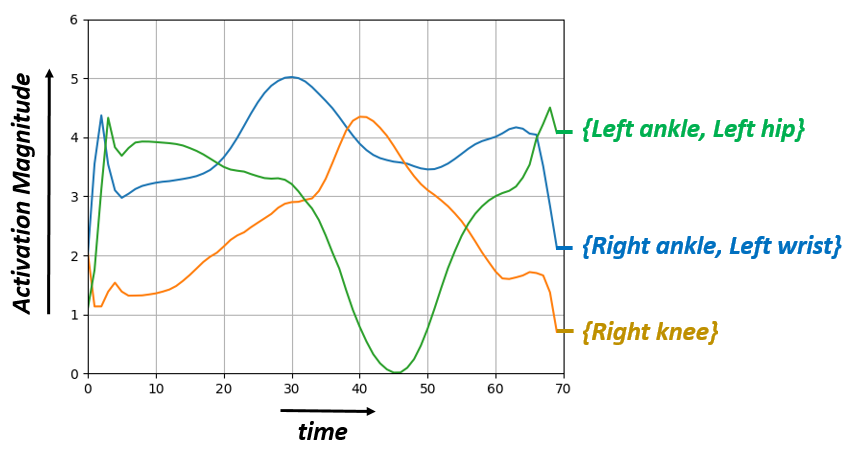}
\end{center}
 \caption{The plot depicts the changes in activation magnitudes of three semantically meaningful filters over time.}
\label{fig:plots}
\end{figure}

The bit about the \textit{swinging motion} can be inferred from the relative change in the magnitude of the activation in the dimension corresponding to the blue filter. Figure \ref{fig:plots} zooms into this particular set of filters and shows their activation magnitudes over the entire video sequence. What is very interesting here is that the activation of the filter corresponding to left ankle and hip joints is close to zero at the peak of the kicking motion. At approximately the same time step, the activation magnitude of the dimension corresponding to the right knee joint peaks. The story that the filters are explaining makes sense. The sequence description that we can interpret from the filters and their activations provides insight into why the model arrived to a certain prediction. During a kicking activity, we first step towards the target with our pivot foot, firmly plant the pivot foot (in this case, the left foot), swing the kicking foot around and step back to return to original position. Note that we focused our analysis on selected interesting dimensions of the hidden representation with significant weight magnitudes. It is important to note that all other positive dimensions also factor into the final decision of the classifier but our discussion was focused on the significant and interesting ones.

\subsection{Comparison to Other State-of-the-Art}
We focused most of our narrative on how a Res-TCN formulation yields explainable spatio-temporal representation compared to state-of-the-art LSTM-based RNN counterparts. We also validate the effectiveness of our model on producing discriminative spatio-temporal features for 3D human action analysis. We compare the performances of published methods on NTURGB+D dataset and show that we improve on the current state-of-the-art on both Cross-Subject and Cross-View settings.

\begin{table}[h!]
\centering
\caption{Comparison to other learning based methods on NTURGB+D skeleton dataset with Cross-Subject (CS) and Cross-View (CV) settings in accuracy (\%).}

\begin{tabular}{|c | c | c |} 
 \hline
 Methods & CS & CV  \\ [0.5ex] 
 \hline

 Dynamic Skeletons \cite{dynamic2015} & 60.2 & 65.2  \\
 HBRNN \cite{du2015} & 59.1 & 64.0  \\
 Deep LSTM \cite{nturgdb2016} & 60.7 & 67.3  \\
 P-LSTM \cite{nturgdb2016} & 62.9 & 70.3  \\
 Trust Gate \cite{Liu2016} & 69.2 & 77.7  \\
 STA-LSTM \cite{Song2017} & 73.4 & 81.2 \\
 \hline\hline
 Res-TCN & \textbf{74.3} & \textbf{83.1}  \\ [1ex] 
 \hline
\end{tabular}

\label{table:1}
\end{table}

\section{Conclusion}
We present a new approach to performing 3D human action analysis with a Res-TCN. We discuss how such an architecture enhances the interpretability of model parameters and features compared to other popular RNN based approaches. Given an interpretable input such as sequence of human skeletons positions, we can begin to explain what each of the learned filters in a Res-TCN are leveraging on to make a prediction. We show that the model learns to pay different levels of attention both spatially and temporally. Experimentally, we validate that our model is explainable and produces a discriminative representation for human activity analysis, improving upon the state-of-the-art. 

{\small
\bibliographystyle{ieee}
\bibliography{egbib}
}

\end{document}